\documentclass[runningheads]{llncs}

\usepackage{eccv}

\usepackage{eccvabbrv}

\usepackage{graphicx}
\usepackage{booktabs}

\usepackage[accsupp]{axessibility}  

\usepackage{multirow}
\usepackage{orcidlink}
\usepackage{bbding}
\usepackage{pifont}
\usepackage{color,xcolor}
\usepackage{colortbl}
\usepackage{multicol}
\usepackage{stfloats}
\usepackage{subcaption}
\usepackage{textcomp}
\usepackage{tikz}
\usepackage{contour}
\usepackage{amsmath}
\usepackage{amssymb}
\usepackage{wasysym}
\usepackage[misc]{ifsym}
\newcommand{\cmark}{\ding{51}\xspace}%
\newcommand{\xmarkg}{\textcolor{lightgray}{\ding{55}}\xspace}%
\newcommand{\pub}[1]{\color{gray}{\scriptsize{[{#1}]}}}
\newcommand{\ours}{SegPoint\xspace}
\newcommand{\ourbaseline}{SegPoint$^\dagger$\xspace}
\newcommand{\ourdataset}{\textit{Instruct3D}\xspace}
\definecolor{lightergray}{gray}{0.93}

\newcommand{\logo}{\makebox[0pt][l]{\hspace{0pt}\raisebox{-0.8ex}{\includegraphics[height=25pt]{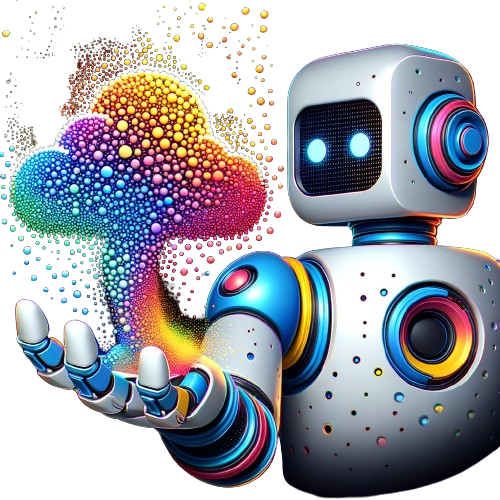}}}}

\usepackage{hyperref}
\newcommand{{\datanum}}{2,565\xspace}
\newcommand{{\datanumtrain}}{2,052\xspace}
\newcommand{{\datanumval}}{513\xspace}
\begin{document}

\title{\logo \ \ \ \ \ \ \ SegPoint: Segment Any Point Cloud via Large Language Model} 

\titlerunning{SegPoint}

\author{Shuting He\inst{1}\orcidlink{0000-0002-1582-5684} \and
Henghui Ding\inst{2}\orcidlink{0000-0003-4868-6526} \and
Xudong Jiang\inst{1}\orcidlink{0000-0002-9104-2315} \and
Bihan Wen\inst{1}\orcidlink{0000-0002-6874-6453}~$^{\textrm{\Letter}}$
}

\authorrunning{S. He et al.}

\institute{Nanyang Technological University \and
Institute of Big Data, Fudan University \\
\email{{\footnotesize\{heshuting555, henghui.ding\}@gmail.com, \{exdjiang, bihan.wen\}@ntu.edu.sg}\\
\url{https://heshuting555.github.io/SegPoint}}}
\footnotetext[0]{${\textrm{\Letter}}$ Corresponding author}
\maketitle

\begin{abstract}
Despite significant progress in 3D point cloud segmentation, existing methods primarily address specific tasks and depend on explicit instructions to identify targets, lacking the capability to infer and understand implicit user intentions in a unified framework. In this work, we propose a model, called \ours, that leverages the reasoning capabilities of a multi-modal Large Language Model (LLM) to produce point-wise segmentation masks across a diverse range of tasks: 1) 3D instruction segmentation, 2) 3D referring segmentation, 3) 3D semantic segmentation, and 4) 3D open-vocabulary semantic segmentation. 
To advance 3D instruction research, we introduce a new benchmark, \ourdataset, designed to evaluate segmentation performance from complex and implicit instructional texts, featuring \datanum point cloud-instruction pairs.
Our experimental results demonstrate that \ours achieves competitive performance on established benchmarks such as ScanRefer for referring segmentation and ScanNet for semantic segmentation, while delivering outstanding outcomes on the \ourdataset dataset. To our knowledge, \ours is the first model to address these varied segmentation tasks within a single framework, achieving satisfactory performance.

\end{abstract}

\begin{figure}[t]
    \centering
    \includegraphics[width=1.\textwidth]{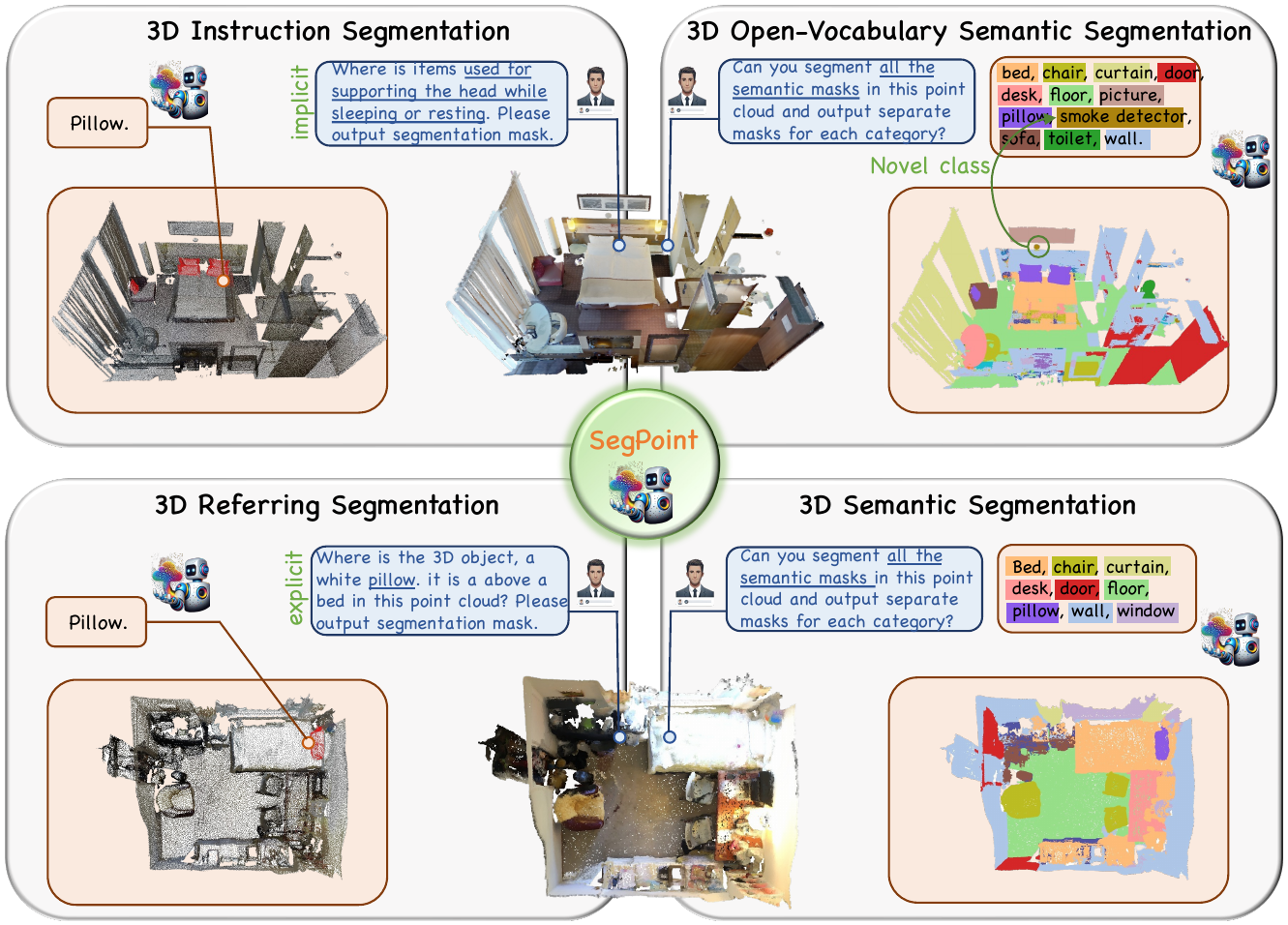} 
    \vspace{-3.36mm}
    \caption{Example of functionality in \ours. \ours can complete various point cloud tasks in a unified framework by leveraging task-specific prompts, including 1) 3D instruction segmentation, 2) 3D referring segmentation, 3) 3D semantic segmentation, and 4) 3D open-vocabulary semantic segmentation.}
    \label{fig:teaser}
    \vspace{-3.6mm}
\end{figure}

\section{Introduction}

3D point cloud segmentation, a critical challenge in the 3D vision community, aims to interpret and classify each point in a point cloud to understand its semantic properties~\cite{pointnet,pointnet++,pointtransformerv1,mask3d,oneformer3d,spformer}. This longstanding issue has spurred significant advancements across various fields, including robotics, autonomous driving, virtual reality, \etc. This challenge has evolved into a series of specialized tasks, each targeting a specific segmentation aspect. Overall, tasks cover basic semantic and instance segmentation~\cite{s3dis,scannet,scannet200,scannet++}, as well as more practical tasks such as referring segmentation~\cite{scanrefer,GRES,referit3d,Multi3drefer,sceneverse,3d-vista}, which segments points based on explicit textual descriptions, and open-vocabulary segmentation~\cite{Openscene,PLA,openmask3d,PMOSR,open3dis,OpenVocabulary} designed for the dynamic and complex nature of real-world.

Despite significant progress achieved within the 3D community toward accurately segmenting objects through specifically designed models, each model is typically developed to tackle one specific segmentation task, leading to inefficiencies and a lack of versatility for real-world application. Furthermore, previous perception approaches heavily depend on predefined categories or explicit expressions for language understanding. Such approaches fall short in interpreting and acting on implicit instructions often found in human language, a critical gap that hinders the development of truly intelligent next-generation perception systems. This brings a pivotal question: \textit{Is it possible to design a unified model capable of comprehensively addressing all aforementioned 3D tasks with human-like instructions?} The exploration of this question not only challenges the current paradigms of 3D point cloud segmentation but also opens the door to groundbreaking advancements in robotic perception and interaction.

In this work, we propose a model called \ours, leveraging the Large Language Model's (LLM) advanced ability to reason and comprehend user instructions. To enhance 3D scene comprehension, we integrate a Geometric Enhancer Module that extracts local semantics from point clouds, seamlessly incorporating this geometric insight into the feature extraction process. Furthermore, a Geometric-guided Feature Propagation is designed to utilize semantic flows derived from geometric priors and the LLM's hidden embeddings, facilitating the generation of fine-grained, high-quality features for accurate dense prediction tasks. Unlike previous attempts at 2D field~\cite{LISA,GlaMM,MeViS}, we do not depend on additional costly pre-trained segmentation models like SAM~\cite{SAM}.

Moreover, we introduce a benchmark named \ourdataset, designed to advance research in the field of segmentation driven by implicit and complex instructions. Understanding these nuanced instructions necessitates reasoning abilities and extensive knowledge of the world.
It includes a total of \datanum diverse pairs of instructions and point clouds for tuning and evaluation. Our comprehensive experiments demonstrate the benchmark's utility in evaluating the model's capability of segmentation based on human-like instructions.

Taking advantage of a multi-modal LLM and task-specific prompts, \ours is capable of generating segmentation masks for a wide range of tasks in a unified model: 1) 3D instruction segmentation, 2) 3D referring segmentation, 3) 3D semantic segmentation, and 4) 3D open-vocabulary semantic segmentation, as depicted in \figurename~\ref{fig:teaser}. \ours achieves competitive results on established benchmarks like ScanRefer~\cite{scanrefer} for referring segmentation and ScanNet~\cite{scannet} for semantic segmentation while showing remarkable performance on the \ourdataset dataset.

In summary, our main contributions are as follows:
\begin{itemize}
\setlength\itemsep{0.1em}

\item{We propose \ours, the first 3D segmentation model that can comprehend human intentions and address multiple segmentation tasks within one framework by harnessing the Large Language Model's reasoning capabilities.}

\item{We present a Geometric Enhancer Module that integrates comprehensive scene information into the process of 3D scene understanding. Besides, A Geometric-guided Feature Propagation is designed to achieve accurate and fine-grained segmentation. These two modules supplement the missing local information and grasp fine-grained features for dense prediction tasks.}

\item{We introduce a new task called 3D instruction segmentation and construct a new dataset \ourdataset, which necessitates a model's self-reasoning to interpret implicit instructions for segmenting the target object.}

\item{Our experimental findings reveal that \ours not only competes strongly in 3D semantic, referring, and open-vocabulary semantic segmentation but also excels in 3D instruction segmentation, showcasing its versatility and effectiveness across a spectrum of segmentation challenges.}
\end{itemize}

\section{Related Work}
\subsection{Multi-modal Large Language Model}
Inspired by the exceptional reasoning abilities of Large Language Models, researchers are delving into transferring these capabilities into the vision realm~\cite{LISA,MOSE,primitivenet,shuai2024survey}, developing multi-modal LLMs. Flamingo~\cite{alayrac2022flamingo}, BLIP-2 \cite{li2023blip}, mPLUG-OWL \cite{ye2023mplug}, Otter~\cite{li2023otter}, LLaVA~\cite{liu2023visual} and MiniGPT-4~\cite{zhu2023minigpt} first construct image-text feature alignment followed by instruction tuning and achieve superior performance. 
Recent studies have increasingly concentrated on integrating foundational models with tasks that demand a refined understanding at the region or pixel level.
VisionLLM~\cite{wang2023visionllm} introduces a versatile interaction interface for various vision-centric tasks via instruction tuning, albeit without fully leveraging LLMs for intricate reasoning tasks. Kosmos-2~\cite{peng2023kosmos} has built substantial data of grounded image-text pairs, thereby embedding grounding capabilities into LLMs. DetGPT~\cite{detgpt} links the fixed multi-modal LLM with an open-vocabulary detector, facilitating user instruction-based detection tasks. This growing interest has spurred further innovations, including GPT4RoI~\cite{zhang2023gpt4roi}, LLaVA-grounding~\cite{LLaVA-grounding}, Ferret~\cite{Ferret}, LISA~\cite{LISA}, GlaMM~\cite{GlaMM}, PixelLM~\cite{PixelLM}, Sphinx~\cite{Sphinx}, each contributing to the evolving landscape of instruction-based, multi-modal understanding.

Building on advancements of multi-modal large language models in 2D image domain, the field is witnessing a seamless transition into 3D spaces.
PointLLM~\cite{pointllm} harnesses the prowess of large language models and trains with 3D point clouds following the paradigm of LLaVA~\cite{liu2023visual}.
3D-LLM~\cite{3dllm} utilizes 2D foundation models to encode multi-view images of 3D point clouds.
Point-Bind~\cite{pointbind} aligns point clouds with Image-Bind~\cite{imagebind} and leverages ImageBind-LLM to reason multi-modality input without 3D-instruction data training.
GPT4Point~\cite{GPT4Point} pioneers in facilitating a unified approach towards 3D object understanding and generation, setting a new standard for versatility.
However, these models primarily concentrate on scene-level insights, often overlooking the intricate details at the region, or point level.
Contrasting with existing models, our research focuses on two pivotal goals: 1) efficiently inject segmentation capabilities into multi-modal LLMs to conduct point-level understanding and 2) design a unified framework for 3D point cloud segmentation via the reasoning ability of LLMs.

\subsection{3D Point Cloud Segmentation}

3D point cloud segmentation, a crucial task in computer vision, can be categorized into 3D semantic, instance, and panoptic segmentation. 3D semantic segmentation~\cite{s3dis,scannet,scannet200,scannet++,pointnet,pointtransformerv1,PAPFZ} assigns each point in a 3D space to specific, predefined classes.
In contrast, 3D instance segmentation~\cite{pointgroup,mask3d,oneformer3d} goes a step further by classifying each point and differentiating between distinct objects of the same class. 
3D panoptic segmentation~\cite{xiao2023position,zhou2021panoptic} aims to group 3D points according to their semantics and identities.
Lately, more practical tasks have emerged, such as 3D referring segmentation~\cite{scanrefer,referit3d,Multi3drefer,butd-detr,EDA,DsHmp}, which extends referring expression segmentation~\cite{vltpami,GRES,MeViS,ding2021vision,M3Att,ISFP} to 3D and segments a target instance based on explicit linguistic descriptions, and 3D open-vocabulary segmentation~\cite{open3dis,openmask3d,Openscene,PLA}, designed to identify and segment unseen objects beyond a fixed set of known categories.

Despite the significant progress made by transformer-based models: Mask3D\\~\cite{mask3d}, SPFormer~\cite{spformer}, MAFT~\cite{MAFT}, and OneFormer3D~\cite{oneformer3d} in basic 3D segmentation tasks, their application in real-world scenarios requiring human language interaction remains constrained. Besides, Seal~\cite{seal} aims to segment any point cloud through distilling vision foundation models, while it doesn't use language as cues.
TGNN~\cite{TGNN} is the first work to tackle referring segmentation problem that proposes aggregating textual features by considering the neighboring local structure of each instance but it heavily depends on explicit expressions or predefined categories for language understanding. 
Furthermore, the development of models tailored to specific segmentation tasks restricts their versatility and applicability in diverse real-world scenarios.
Therefore, it is imperative to develop more intelligent interaction ways and a unified model for 3D point segmentation.

\section{Approach}
\begin{figure}[t]
    \centering
    \includegraphics[width=1.\textwidth]{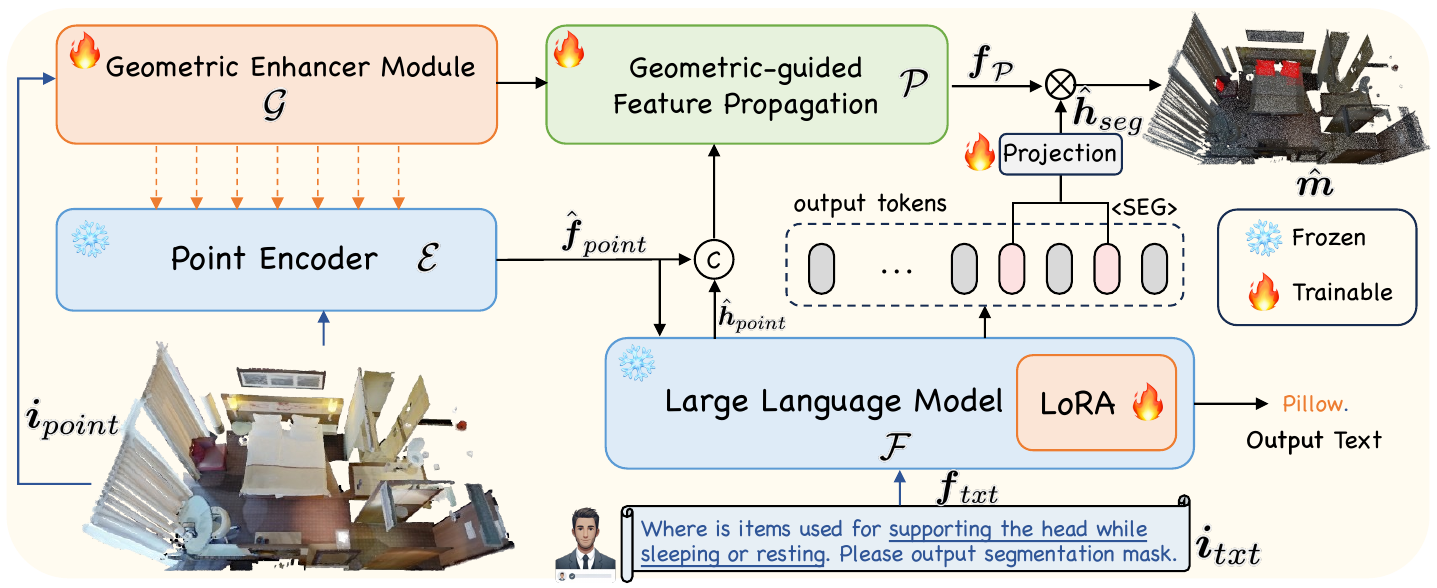} 
    \vspace{-0.5cm}
    \caption{The pipeline of \ours. Given input point cloud and text query, the multi-modal LLM $\mathcal{F}$ generates text output. Geometric Enhancer Module $\mathcal{G}$ injects geometric information into Point Encoder $\mathcal{E}$ and obtains point features $\hat{\vec{f}}_{point}$. Per-point embeddings ${\vec{f}}_{\mathcal{P}}$ derived from Geometric-guided Feature Propagation $\mathcal{P}$ multiplied with the embedding associated with the \texttt{<SEG>} token yield the final segmentation masks.}
    \label{fig:framework}
    \vspace{-0.4cm}
\end{figure}

\subsection{Architecture Overview}
The overall architecture of \ours is presented in \figurename~\ref{fig:framework}. \ours mainly comprises four parts: \textit{i}) a pre-trained point encoder $\mathcal{E}$ tailored for aligning with textual data; \textit{ii}) a large language model $\mathcal{F}$ endowed with advanced reasoning capabilities; \textit{iii}) a Geometric Enhancer Module $\mathcal{G}$ responsible for extracting geometric representation from input point clouds and infusing these priors into the point encoder; and \textit{iv}) a Geometric-guided Feature Propagation $\mathcal{P}$ which is key to achieving precise mask generation. The collaboration between the Geometric Enhancer Module and Geometric-guided Feature Propagation is crucial, as it equips LLMs with the ability to generate masks effectively in various scenarios.
    \vspace{-0.2cm}
\subsection{Vanilla Baseline}\label{sec:baseline}

The input of the framework is the text instructions $\vec{i}_{txt}$ and point cloud $\vec{i}_{point}\in \mathbb{R}^{N\times(3+F)}$. Specifically, a point cloud scene, comprising $N$ points, each includes 3D coordinates $\in \mathbb{R}^3$ and an auxiliary feature vector $\in \mathbb{R}^F$ (e.g., color). The point cloud $\vec{i}_{point}$ is fed into the point encoder $\mathcal{E}$, which extracts point features $\vec{f}_{point} \in \mathbb{R}^{N_1 \times D}$, where $N_1 \ll N$, $D$ is the feature dimension. Concurrently, the text instruction $\vec{i}_{txt}$ undergoes tokenization via $\mathcal{F}_{tokenize}$. These prepared inputs are then fed into the Large Language Model $\mathcal{F}$, resulting in a textual response ${\vec{y}}$. The above process can be formulated as:
\begin{equation}
\vec{f}_{point} = \mathcal{E}(\vec{i}_{point}), \quad  \vec{f}_{txt} = \mathcal{F}_{tokenize}(\vec{i}_{txt}), \quad 
    {\vec{y}} = \mathcal{F}(\vec{f}_{point}, \vec{f}_{txt}).
\end{equation}

Building on the approach introduced by LISA~\cite{LISA}, \ours enhances the segmentation capabilities of Large Language Models (LLMs) by expanding their vocabulary with a new special token, \texttt{<SEG>}. This modification enables the model to recognize and predict the \texttt{<SEG>} token within the output sequence as a signal to identify segmentation targets. Upon detecting a \texttt{<SEG>} token, the corresponding output sequence belonging to \texttt{<SEG>} token is extracted and processed through an MLP layer $\gamma$, generating mask embeddings $\vec{h}_{seg}$. The final step involves computing each binary mask prediction $\vec{m} \in \mathbb{R}^{N}$ by performing a dot product between the mask embeddings $\vec{h}_{seg}$ and the upsampled per-point embeddings derived from the point features $\vec{f}_{point}$.  The formulation of the aforementioned process is given by:
\begin{align}
\begin{aligned}
    \vec{h}_{seg} = \gamma({\vec{y}}_{[seg]} &), \quad \vec{m} = \vec{h}_{seg} \otimes {UpS.} (\vec{f}_{point}), \\
\end{aligned}
\end{align}
where ${UpS.}$ denotes the upsampling operation following PointNet++~\cite{pointnet++} on $\vec{f}_{point}$. 
The vanilla baseline represents an initial attempt to bridge the gap between LLMs' text comprehension and point cloud segmentation tasks. It encounters two primary issues. Firstly, the point encoder is trained on a scene-level dataset for classification achieving alignment between text and point clouds, not specifically trained for dense prediction tasks. Besides, the point encoder's first layer employs Farthest Point Sampling (FPS)~\cite{pointnet++} to reduce the point cloud to $N_1$ points, risking the loss of details vital for accurate dense predictions. Secondly, the operation of directly upsampling from $N_1$ to $N$ points to obtain per-point embeddings is prone to losing structural information and introducing a notable degree of noise, undermining the model's efficacy in segmentation tasks.

\subsection{Geometric Enhancer Module}

\begin{figure}[t]
    \centering
    \includegraphics[width=1.\textwidth]{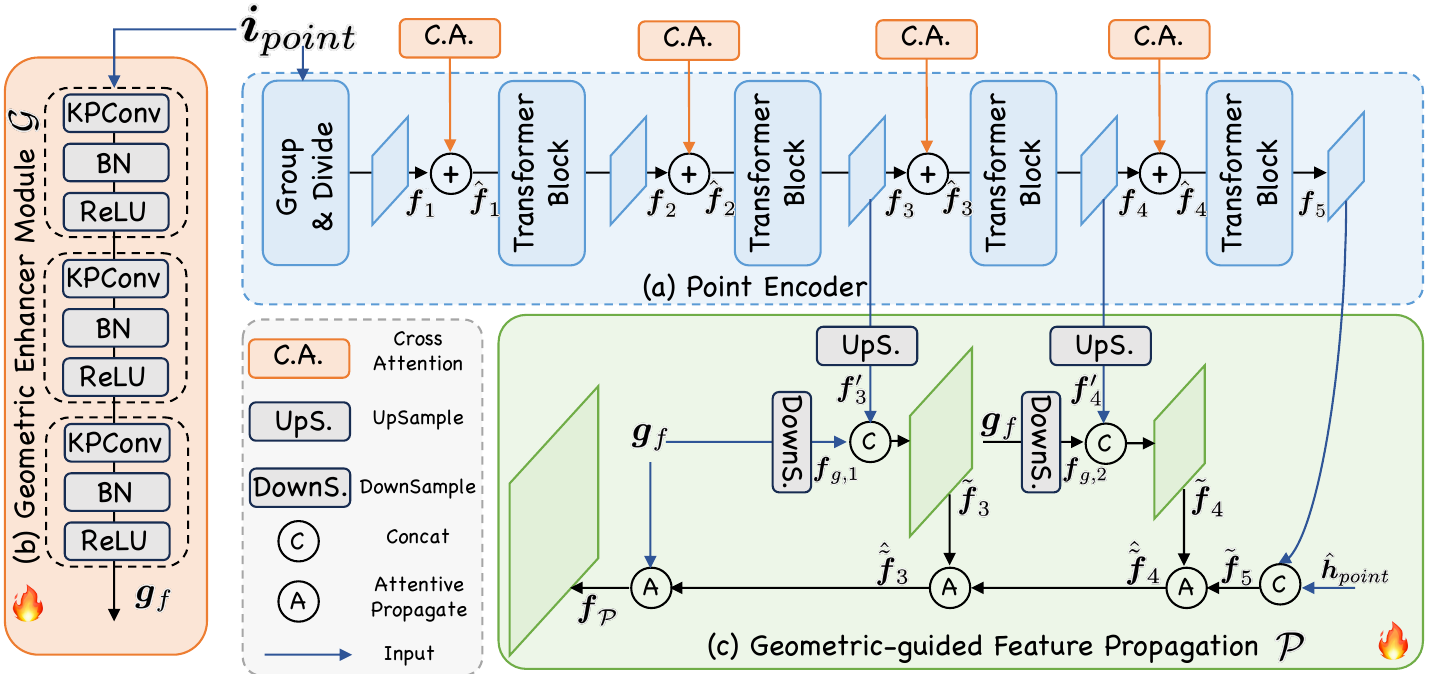} 
    \vspace{-4.36mm}
    \caption{Architecture of the proposed (b) Geometric Enhance Module (GEM) and (c) Geometric-guided Feature Propagation (GFP) interaction with (a) Point Encoder.}
    \label{fig:mask_decoder}
    \vspace{-2.6mm}
\end{figure}

To adapt the pre-trained point encoder for dense prediction tasks while maintaining its superior scene recognition capability, our objective is to harness the geometric information across the entire scene to guide the further feature learning process. Drawing inspiration from recent advancements in 2D computer vision, where studies~\cite{pvtv2,cvt,vitadapter,park2022vision} demonstrate that convolutions enhance transformers' ability to capture local spatial information, we introduce the Geometric Enhancer Module (GEM). This module is specifically designed to grasp the local geometric contexts within point clouds while enabling the preservation of the point encoder's foundational architecture and information integrity.

As shown in \figurename~\ref{fig:mask_decoder}, Geometric Enhancer Module $\mathcal{G}$ is composed of three blocks, each featuring a KPConv~\cite{kpconv} layer followed by BN and ReLU activation. The architecture is similar to the 2D convolutional stem~\cite{resnet}. 
We utilize KPConv instead of vanilla convolution or linear layer here to facilitate grasping the local geometric information effectively. 
The resultant geometric feature, represented by $\vec{g}_{f} \in \mathbb{R}^{N \times D}$, contains the features across all points, thereby supplementing the missing local information. 
This $\vec{g}_{f}$ is then leveraged to infuse geometric insights into the point encoder's features via a cross-attention mechanism, the above process can be expressed as:
\begin{equation}
 \vec{g}_{f} = \mathcal{G}(\vec{i}_{point}), \quad  \hat{\vec{f}_i} =  \vec{f}_i+g_i \cdot \mathrm{softmax}\left(\frac{\vec{f}_i \vec{g}_{f}^{T}}{\sqrt{D}}\right)\vec{g}_{f},
\end{equation}
where $\vec{f}_i$ represents the feature from the $i$-th block of the point encoder and $l$ consecutive transformer layers are regarded as one block for the convenience of explanation. To fine-tune the integration of geometric information, we introduce a learnable gating factor $g_i$ that modulates the balance between the attention layer's output and the input feature $\vec{f}_i$. This gating factor is initially set to zero, to ensure that the incorporation of geometric data does not abruptly alter $\vec{f}$ feature distribution. Such an approach allows for the preservation and effective utilization of the point encoder's pre-trained weights.
Upon processing through the Geometric Enhancer Module (GEM), the modified output of the point encoder, LLM are formulated as:
\begin{align}
\begin{aligned}
    \hat{\vec{f}}_{point} = \mathcal{E}(\vec{i}_{point}, \vec{g}_{f}),\quad 
    {\hat{\vec{y}}} = \mathcal{F}(\hat{\vec{f}}_{point}, \vec{f}_{txt}),\quad  \hat{\vec{h}}_{seg} = \gamma(\hat{\vec{y}}_{[seg]} &).
\end{aligned}
\end{align}

\subsection{Geometric-guided Feature Propagation}

Addressing the challenge of upsampling point clouds from a sparse set of $N_1$ points to a denser set of $N$ points is crucial, as direct upsampling inevitably introduces noise and results in information loss, leading to sub-optimal performance in segmentation tasks. To mitigate these issues, we introduce Geometric-guided Feature Propagation designed to generate high-quality per-point embeddings. Geometric features $\vec{g}_f$, which carry comprehensive point information, serving as a ``gold message'' for enhancing the upsampling process. By integrating these geometric features, we aim to significantly improve the quality and accuracy of the generated dense per-point embeddings.

As illustrated in \figurename~\ref{fig:mask_decoder}, we begin by upsampling the higher-layer features ${\vec{f}_3,\vec{f}_4}$ from a smaller set of points $N_1$ to larger sets $N_3, N_2$, employing PointNet++'s~\cite{pointnet++} propagation techniques. This step yields features ${\vec{f}'_3 \in \mathbb{R}^{N_3\times D}}$, and ${\vec{f}'_4 \in \mathbb{R}^{N_2\times D}}$.
Subsequently, we perform downsampling on the geometric features $\vec{g}_f$ 
from the original number of points $N$ to reduced counts $N_2, N_3$, respectively, utilizing the Farthest Point Sampling (FPS) technique. In this process, we directly obtain the features of the sampled points without performing additional $k$-nearest neighbor ($k$-NN) and pooling operations to simplify the computation and produce features ${\vec{f}_{g,1} \in \mathbb{R}^{N_3\times D}}$, and $\vec{f}_{g,2} \in \mathbb{R}^{N_2\times D}$.

In the next phase, we integrate the up- and downsampled features, processing them through fully connected layers and ReLU activation to update the feature vectors ${\Tilde{\vec{f}}_{3} \in \mathbb{R}^{N_3\times D}}$, and $\Tilde{\vec{f}}_{4} \in \mathbb{R}^{N_2\times D}$. Note that the last layer feature ${\vec{f}}_{5}$ bypasses this step. Instead, we concatenate it with $\hat{\vec{h}}_{point}$ from the LLM output to form $\Tilde{\vec{f}}_{5}$ to perceive multi-modal information from LLM.

Finally, to enable information exchange across different point densities, we propose attentive propagation. Take the propagation from $\Tilde{\vec{f}}_{5}$ to ${\Tilde{\vec{f}}_{4}}$ as example. Here, $\Tilde{\vec{f}}_{4} \in \mathbb{R}^{N_2\times D}$ acts as a set of local centers. For each local center within $\Tilde{\vec{f}}_{4}$, we identify its neighboring points from $\Tilde{\vec{f}}_{5}$ using the $k$-NN algorithm, resulting in $\vec{f}_{54} \in \mathbb{R}^{N_2 \times k \times D}$. 
Then, employing cross-attention mechanism, where $\Tilde{\vec{f}}_{4}$ serves as query and $\vec{f}_{54}$ as both key and value, facilitates information flow across different point densities and effectively extract relevant details into the query points.
\begin{equation}\label{eq:Qs}
\hat{\Tilde{\vec{f}}}_{4} =  \Tilde{\vec{f}}_{4}+\mathrm{softmax}\left(\frac{\Tilde{\vec{f}}_{4} \vec{f}_{54}^{T}}{\sqrt{D}}\right)\vec{f}_{54}.
\end{equation}

Leveraging geometric-guided feature propagation enables us to produce high-quality per-point embeddings denoted as ${\vec{f}}_{\mathcal{P}}$, laying the foundation for generating precise segmentation masks expressed as follows:
\begin{align}
\begin{aligned}
    {\vec{f}}_{\mathcal{P}} = \mathcal{P}(\hat{\vec{f}}_{point}, \vec{g}_{f}), \quad \hat{\vec{m}} = \hat{\vec{h}}_{seg} \otimes {\vec{f}}_{\mathcal{P}} . \\
\end{aligned}
\end{align}

\subsection{Training Objectives}
Our model is trained end-to-end leveraging the text classification loss and the segmentation mask loss:
\begin{equation}
    \mathcal{L} = \lambda_{txt} \mathcal{L}_{txt} + \lambda_{bce} \mathcal{L}_{bce}+ \lambda_{dice} \mathcal{L}_{dice},
\end{equation}
where $\mathcal{L}_{txt}$ denotes the auto-regressive cross-entropy loss targeting text generation accuracy, segmentation mask loss includes both binary cross-entropy (BCE) loss $\mathcal{L}_{bce}$ and DICE loss $\mathcal{L}_{dice}$, aims at refining segmentation quality. The weights $\lambda_{txt}$, $\lambda_{bce}$ and $\lambda_{dice}$ are utilized to balance the different loss items. The model's training is guided by the ground-truth labels $\vec{y}_{txt}$ for text and $\vec{M}$ for masks.

\subsection{\ourdataset Dataset Collection}

Although 3D instruction segmentation and 3D referring segmentation~\cite{scanrefer,referit3d,Multi3drefer} are both language-based segmentation,
3D referring segmentation guides segmentation with explicit target object names, e.g., ``chair'', lacking more complicated reasoning instructions, e.g., ``Where to sit in the room?''.
Besides, they also fall short in offering multi-target question-answer pairs with target descriptions directly connected to multiple segmentation masks, which cannot meet a common requirement in real-world scenarios, like ``How to play computer games''.

To enhance the assessment and analysis of instruction segmentation capabilities, we have developed a benchmark, referred to \ourdataset. This benchmark incorporates 280 scenes specifically selected for instruction segmentation tuning and evaluation, sourced from the recently introduced ScanNet++~\cite{scannet++} dataset. Each scene comes with approximately 10 different segmentation instructions, resulting in \datanum instruction-point cloud pairings. 
This dataset is then divided into two splits: {\tt train}, and {\tt val}, containing \datanumtrain, and \datanumval question-answer pairs, respectively. 
Our dataset is uniquely designed to encompass both multi-target and zero-target scenarios, addressing the real-world requirement of identifying multiple objects in response to text queries and accounting for situations where objects mentioned in the text may not be present in the paired point cloud. Besides, we take into account the characteristics of 3D scenes and incorporate diverse locations and view descriptions e.g., ``something that is used for sitting while working at a desk. It is the one facing the window.''. The model needs to have not only reasoning capabilities but also the ability to perceive views and directions in 3D scenes.
These designs underscore the dataset's practical value.

\section{Experiments}
\subsection{Datasets and Evaluation Metrics}\label{sec:dataset_analysis}
\vspace{-1mm}
\label{exp:setting}

\noindent\textbf{Datasets}.
Our training data is composed of two types of datasets: (1) semantic segmentation dataset including ScanNet200~\cite{scannet200}, and S3DIS~\cite{s3dis}; (2) referring segmentation dataset consisting of ScanRefer~\cite{scanrefer}, ReferIt3D~\cite{referit3d}(including Sr3D and Nr3D), and Multi3DRefer~\cite{Multi3drefer}. We design task-specific prompts to facilitate the joint training of various tasks within a unified framework.

\noindent\textbf{Semantic Segmentation Dataset.} 
We use two strategies to generate templates. 1) segment the specific category: ``\texttt{\textbf{USER}}: \texttt{<POINT>} \texttt{\small Can you segment the} \texttt{\{category\}} \texttt{\small \underline{category} in this point cloud?} \texttt{\textbf{ASSISTANT}}: \texttt{\{category\}} \texttt{<SEG>}.'', where \texttt{category} is the random chosen category, and \texttt{<POINT>} denotes the placeholder for tokens of point cloud patches. 2) segment all the categories: ``\texttt{\textbf{USER}}: \texttt{<POINT>} \texttt{\small Can you segment all the semantic masks in this point cloud and \\output separate masks for each category in the alphabetical order of the\\ categories?} \texttt{\textbf{ASSISTANT}}: \texttt{\{category\}} \texttt{<SEG>, \texttt{\{category\}} \texttt{<SEG>, ...}}'' To simplify the output and ensure it has only one possible answer, we add the constraints ``\texttt{\small in the alphabetical order of the categories}''. 
To avoid generating class names not in the dataset, we incorporate category names in a dataset into the prompts during training and inference.

\noindent\textbf{Referring Segmentation Dataset.} 
We use template prompts: ``\texttt{\textbf{USER}}: \texttt{<POINT>}  \texttt{\small Can you segment the \underline{object}} \texttt{\{description\}} \texttt{\small in this point cloud?} \texttt{\textbf{ASSISTANT}}: \texttt{\{category\}} \texttt{<SEG>}.'', where \texttt{\{description\}} is the given explicit description from referring segmentation dataset. 
It is worth noting that during training, we also use other templates to generate the QA data to ensure data diversity. We add \texttt{\{category\}} in front of \texttt{<SEG>} to unify the output format so that when outputting semantic masks, the output category name is the label it predicts.

\noindent\textbf{Evaluation Metrics}. We follow most previous works on 3D segmentation~\cite{mask3d,oneformer3d,TGNN} to adopt mIoU as primary metric. mIoU is defined by the average of all per-point cloud scene Intersection-over-Unions (IoUs). Besides, we employ accuracy (Acc) as a metric to evaluate whether the model accurately identifies targets with which the predictions have an IoU greater than 0.5.

\vspace{-4mm}
\subsection{Implementation Details}
\vspace{-1mm}

In our experiments, unless specified otherwise, we employ the LLaMA2-7B model\\~\cite{llama} as the large language model $\mathcal{F}$ and Uni3D~\cite{uni3d} as the point cloud processing backbone $\mathcal{E}$. 
The training stage leverages the deepspeed~\cite{rasley2020deepspeed} engine for efficiency, with the AdamW~\cite{loshchilov2017decoupled} optimizer guiding the learning process. The learning rate and weight decay are set to $0.0003$ and $0$, respectively, enhanced by a WarmupDecayLR learning rate scheduler that initiates with $100$ warmup iterations. The projection layer $\gamma$ utilizes an MLP with channel sizes of [256, 4096, 4096]. We set balancing weight $\lambda_{txt\_gen}$, $\lambda_{bce}$, and $\lambda_{dice}$ to 1.0, 2.0, 2.0, respectively. The experiments utilize a total batch size of 16, distributed across 4 NVIDIA 80G A100 GPUs, and span 5,000 iterations, culminating in a training period of approximately 3 days. During training, we use all mentioned datasets in Sec.~\ref{sec:dataset_analysis} for joint training by leveraging task-specific prompts. 
For evaluation on a specific dataset, we finetune the trained model on the corresponding dataset.

\vspace{-4mm}
\subsection{Results on \ourdataset}\label{sec:instruct3d}
\vspace{-1mm}
The instruction segmentation results, as detailed in Table~\ref{tab:reason_seg}, underscore a significant advancement: where existing methodologies fall short, our model demonstrates exceptional prowess, achieving a more than 15\% improvement in mIoU for tasks requiring intricate reasoning. Unlike conventional referring segmentation tasks, instruction segmentation demands not just identification but also understanding, necessitating the model's reasoning capabilities and access to world knowledge. Existing approaches, confined to explicit references, struggle with implicit queries due to their lack of understanding, which further underscores the task's inherent challenges. In contrast, our model leverages LLMs to bridge this gap, demonstrating superior performance by comprehending and interpreting the queries accurately. 
Moreover, \ours configuration substantially outperforms \ourbaseline, highlighting the critical role of our designed Geometric Enhancer Module and Geometric-guided Feature Propagation components. Notably, even in its baseline form, \ourbaseline surpasses all competing methods, validating the effectiveness and rationale behind our pipeline design.

Besides, different from traditional two-stage approaches that first generate mask proposals using a pre-trained segmentor like Mask3D~\cite{mask3d} and then apply language-aware networks for selection, \ours directly tackles the task, bypassing the need for preliminary mask proposals, enhancing its efficiency.

\begin{table*}[t]
\scriptsize
\caption{\textbf{3D instruction segmentation benchmark results} on \ourdataset val split evaluated by Acc and mIoU. $\dagger$ denotes our vanilla baseline removing geometric enhancer module and geometric-guided feature propagation. $*$ represents adding an auxiliary mask head through our implementation.
} 
\vspace{-3mm}
\centering
\setlength{\tabcolsep}{16pt}
\renewcommand{\arraystretch}{1.1}
\begin{tabular}{c|l|c|c|c}
\hline
\rowcolor[gray]{.92}
Stage& Method&   Reference& Acc & mIoU \\

\hline
Two&ScanRefer~\cite{scanrefer}& \pub{ECCV'20}&12.0 & 6.9 \\
Two&ReferIt3D~\cite{referit3d}& \pub{ECCV'20} &11.7 & 6.4\\
Two&M3DRef-CLIP~\cite{Multi3drefer}& \pub{ICCV'23}&18.1 & 12.8 \\
\hline
Single &TGNN~\cite{TGNN}      & \pub{AAAI'21} &12.9 & 7.1  \\ 
Single&BUTD-DETR~\cite{butd-detr}$^*$  & \pub{ECCV'22} &16.3 & 10.9 \\
Single&EDA~\cite{EDA}$^*$  & \pub{CVPR'23}      &16.6 & 12.1    \\
\hline
\rowcolor{eccvblue!10} Single&\ourbaseline      &  \pub{ECCV'24}  &  \textbf{21.8}  & \textbf{16.1} \\
\rowcolor{eccvblue!10}Single&{\ours}      & \pub{ECCV'24} &\textbf{31.6}  &\textbf{27.5}     \\
\hline
\end{tabular}

\label{tab:reason_seg}
\end{table*}

\vspace{-4mm}
\subsection{Results on Semantic Segmentation}
\vspace{-1mm}

Table~\ref{tab:semantic_seg} present \ours's performance on semantic segmentation, delivering competitive results across diverse datasets. Our model uses a simple yet effective answer format, \texttt{{category}} \texttt{<SEG>}, to use \texttt{{category}} name as predicted labels, achieving particularly stronger performance in datasets with various categories like ScanNet200~\cite{scannet200}, where it surpasses SOTA methods by 2.1\% mIoU. To ensure fair comparisons, we fine-tune our model on each semantic segmentation dataset to accommodate the varying class category definitions.

\begin{table}[t]
\scriptsize
\caption{\textbf{3D Semantic segmentation benchmark results} on S3DIS~\cite{s3dis}, ScanNet~\cite{scannet}, and ScanNet200~\cite{scannet200}. We evaluate on the Area 5 of S3DIS and validation split of ScanNet and ScanNet200.}
\vspace{-3mm}
  \renewcommand{\arraystretch}{1.1}
  \setlength{\tabcolsep}{12pt}
  \centering
  \begin{tabular}{l|c|c|c|c}
  \hline
  \rowcolor[gray]{.92}
  Method& Reference  &  {ScanNet} & {ScanNet200} & {S3DIS} \\
  \hline
PointNet++~\cite{pointnet++} &\pub{NeurIPS'17}& 53.5 & - &- \\
MinkUNet~\cite{minkowski} &\pub{CVPR'19} &72.2 &25.0 &65.4 \\
PTv1~\cite{pointtransformerv1} &\pub{ICCV'21}& 70.6  &27.8 &70.4\\
PTv2~\cite{pointtransformerv2} &\pub{NeurIPS'22}& 75.4 &30.2  &71.6 \\
PointNeXt~\cite{pointnext} &\pub{NeurIPS'22}& 71.5  &-  &70.5  \\
OctFormer~\cite{wang2023octformer} &\pub{SIGGRAPH'23}& \textbf{75.7} &32.6  &-  \\
Swin3D~\cite{swin3d} &\pub{ArXiv}& 75.5  &-  &\textbf{72.5} \\
\hline
\rowcolor{eccvblue!10}{\ours}  &\pub{ECCV'24}    & 74.1   & \textbf{35.3}   & 72.4     \\
\hline
  \end{tabular}
  \vspace{-5mm}
\label{tab:semantic_seg}
\end{table}

\vspace{-4mm}
\subsection{Results on Referring Segmentation}
\vspace{-1mm}

\begin{table*}[t]
\scriptsize
\centering
\caption{\textbf{3D referring segmentation benchmark results} on ScanRefer~\cite{scanrefer}, Nr3D~\cite{referit3d}, and Multi3Drefer~\cite{Multi3drefer} evaluated by mIoU. $^*$ represents adding an auxiliary mask head through our implementation.
} 
\vspace{-3mm}
\setlength{\tabcolsep}{9pt}
\renewcommand{\arraystretch}{1.1}
\begin{tabular}{c|l|c|c|c|c}
\hline
\rowcolor[gray]{.92}
Stage& Method&   Reference& ScanRefer & Nr3D  & Multi3DRefer \\

\hline
Two&M3DRef-CLIP~\cite{Multi3drefer}& \pub{ICCV'23}& 35.7  & 27.0 &32.6 \\
Two&3D-STMN~\cite{3D-STMN} & \pub{AAAI'24}  &39.5 & -&-\\
\hline
Single &TGNN~\cite{TGNN}      & \pub{AAAI'21}  &27.8&- & -    \\ 
Single&BUTD-DETR~\cite{butd-detr}$^*$  & \pub{ECCV'22} &35.4  & 27.5& 26.2 \\
Single&EDA~\cite{EDA}$^*$  & \pub{CVPR'23}      & 36.2  & 29.3 &  28.9  \\
Single&X-RefSeg3D~\cite{X-RefSeg3D} & \pub{AAAI'24}  &29.9 & -& -\\
Single&RefMask3D~\cite{RefMask3D} & \pub{ACMMM'24}  &\textbf{44.8} & -&- \\

\hline
\rowcolor{eccvblue!10}Single&{\ours}   &  \pub{ECCV'24} & {41.7}  &\textbf{32.2}  & \textbf{36.1}       \\
\hline
\end{tabular}

\label{tab:refer_seg}
\end{table*}

Table~\ref{tab:refer_seg} presents results on referring segmentation datasets. \ours showcases outstanding performance in both single-target (e.g., ScanRefer~\cite{scanrefer}, Nr3D~\cite{referit3d}) and multi-target and zero-target contexts within the Multi3DRefer~\cite{Multi3drefer} dataset. For multi-targets, we aggregate masks into a single ground truth, and for zero-target, we use an empty mask, indicated by ``\texttt{\textbf{ASSISTANT}}: There is no mask.'' \ours significantly surpasses other approaches, achieving 2.5\% mIoU increase. The evaluation process of two-stage method M3DRef-CLIP is similar to Sec.\ref{sec:instruct3d}.

\vspace{-4mm}
\subsection{Results on Open-vocabulary Semantic Segmentation}
\vspace{-1mm}
Table~\ref{tab:open_semantic_seg} shows our method's open-vocabulary segmentation performance which is directly evaluated on ScanNet++~\cite{scannet++} following the setting in prevalent methodologies in the 2D domain~\cite{ODISE,maskclip}. It demonstrates our superiority over both existing open-vocabulary techniques and even several supervised approaches, showing our model's robust generalization capabilities. It effectively aligns and interprets categories with visual scenes, underscoring the reasoning prowess of large language models. A notable issue is the potential misalignment between output categories of \ours and val split category names. To address this, we employ GPT-4 to match its most similar category names in the val split.

\begin{table}[t]
\scriptsize
\caption{\textbf{3D open-vocabulary semantic segmentation benchmark results} on val split of ScanNet++~\cite{scannet++}.} 
\vspace{-3mm}
\centering
  \renewcommand{\arraystretch}{1.1}
  \setlength{\tabcolsep}{16pt}
  \begin{tabular}{c|l|c|c}
  \hline
  \rowcolor[gray]{.92}
Type&Method& Reference  &  {ScanNet++}\\
\hline
\multirow{4}{*}{Supervised}&PointNet~\cite{pointnet}&\pub{CVPR'17}&  \textcolor{gray}{7.0}\\
&PointNet++~\cite{pointnet++} &\pub{NeurIPS'17}& \textcolor{gray}{15.0} \\
&MinkUNet~\cite{minkowski}&\pub{CVPR'19}& \textcolor{gray}{28.0}\\
&KPConv~\cite{kpconv}&\pub{ICCV'19}& \textcolor{gray}{30.0}\\
\hline
\multirow{2}{*}{Open-Vocabulary}&OpenScene~\cite{Openscene}&\pub{CVPR'23}&12.8 \\
&PLA~\cite{PLA}&\pub{CVPR'23}&14.2\\
&RegionPLC~\cite{RegionPLC}&\pub{CVPR'24}&14.9\\
\hline
\rowcolor{eccvblue!10} Unified&{\ours}      & \pub{ECCV'24}   & \textbf{19.3}       \\
\hline
  \end{tabular}
\label{tab:open_semantic_seg}
\vspace{-5mm}
\end{table}

\vspace{-1mm}
\begin{table*}[t]
\scriptsize
 
      \caption{
    \textbf{Ablation studies} on \ourdataset and ScanRefer. 
    }
    \begin{subtable}[t]{0.5\linewidth}
      \renewcommand{\arraystretch}{1.1}
        \centering
        \begin{tabular}{c|c c |c |c }
\rowcolor[gray]{.9}
\hline 
&\multicolumn{2}{c|}{Components} & {\ourdataset}& {ScanRefer} \\
\rowcolor[gray]{.9}

Index&GEM & GFP  &  mIoU &  mIoU \\
\hline 
\hline
0& \xmarkg& \xmarkg & 16.1&  {30.3}  \\ 
1& \cmark& \xmarkg & 21.4 & 35.8   \\ 
2& \xmarkg& \cmark & 23.2 &  38.1  \\ 
\rowcolor{eccvblue!10} 3&\cmark & \cmark & \textbf{27.5} & \textbf{41.7} \\
\hline 
\end{tabular}
         \vspace{-0.5mm}
        \caption{Effect of different main components. }
        \label{table:3c}
    \end{subtable}
\hfill
    \begin{subtable}[t]{0.5\linewidth}
      \renewcommand{\arraystretch}{1.1}
        \centering
        \begin{tabular}{l |c |c }
\rowcolor[gray]{.9}
\hline 
Method & {\ourdataset}& {ScanRefer} \\
\hline
\hline
Full tuning & 26.2 &  39.1  \\
LoRA~\cite{LoRA} & 24.8 &  38.5   \\ 
MLP & 23.1 &  36.8 \\
\hline
\rowcolor{eccvblue!10} GEM & \textbf{27.5} & \textbf{41.7}\\
\hline 
\end{tabular}
        \vspace{-0.5mm}
        \caption{Effect of different tuning methods. }
    \end{subtable}
\vspace{-5mm}
     \label{tab:ablation_study}
\end{table*}

\vspace{-4mm}
\subsection{Ablation Study}
\vspace{-1mm}
We conduct extensive experiments to verify the effectiveness of our proposed components in \tablename~\ref{tab:ablation_study} (a) on both \ourdataset and ScanRefer~\cite{scanrefer} dataset. 
We established a vanilla baseline as described in Sec.~\ref{sec:baseline} following the paradigm of LISA~\cite{LISA}, which cannot achieve satisfactory performance and only obtain 16.1\%/30.3 mIoU on \ourdataset/ScanRefer, respectively.
Further analysis, both qualitative and quantitative, of our proposed components reveals that their integration substantially outperforms the baseline.

\noindent\textbf{Geometric Enhancer Module}
Integrating the Geometric Enhancer Module (GEM) into our point encoder results in a notable 5.3\%/5.5\% mIoU improvement on \ourdataset/ScanRefer, effectively addressing compatibility issues with dense prediction tasks. An ablation study, shown in \tablename~\ref{tab:ablation_study} (b), shows that our improvement is not due to an increase in parameters. Our approach outperforms traditional full fine-tuning, LoRA~\cite{LoRA} strategies, and the addition of MLP layers for feature adapter, underscoring its effectiveness in embedding 3D domain-specific knowledge into point encoder. Unlike adapter techniques common in language and 2D image processing, our GEM is designed to address the unique challenges in dense prediction tasks.

\vspace{-1mm}
\begin{figure}[t]
    \centering
	\includegraphics[width=1.0\textwidth]{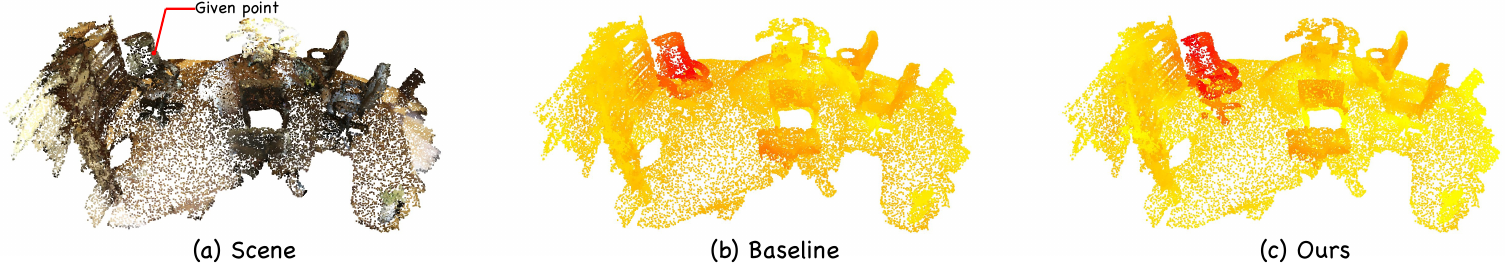}
	\vspace{-0.56cm}
	\caption{(Best viewed in color) We visualize the feature responses between a given point (in \textcolor{red}{red}) and other points in the scene from per-point embeddings ${\vec{f}}_{\mathcal{P}}$ for the baseline and our \ours, 
 respectively. The color changes from yellow to red, indicating increasing feature similarity.} 
	\vspace{-0.3cm}
	\label{fig:feature_visualization}
\end{figure}

\noindent\textbf{Geometric-guided Feature Propagation}
Introducing Geometric-guided Feature Propagation (GFP) results in a substantial improvement over the baseline, as shown in \tablename~\ref{tab:ablation_study} (a) (index 2). This underscores our method's capability to minimize information loss and reduce noise during the upsampling phase, leading to higher-quality per-point embeddings.

\vspace{-4mm}
\subsection{Qualitative Visualization}
\vspace{-1mm}
\figurename~\ref{fig:feature_visualization} qualitatively illustrates the feature responses between a given point and others in a scene. From left to right, it presents the original scene, baseline method, and our \ours, respectively, with warmer colors indicating closer feature relationships. The baseline method predominantly highlights spatially proximate points, often missing important but distant features, such as chair cushions. In contrast, \ours leverages global information, allowing for the identification of both near and distant relevant features, such as cushions and casters. This demonstrates our model's superior ability to capture global context and recognize intricate structures within the scene.

\begin{figure}[t]
    \centering
	\includegraphics[width=1.0\textwidth]{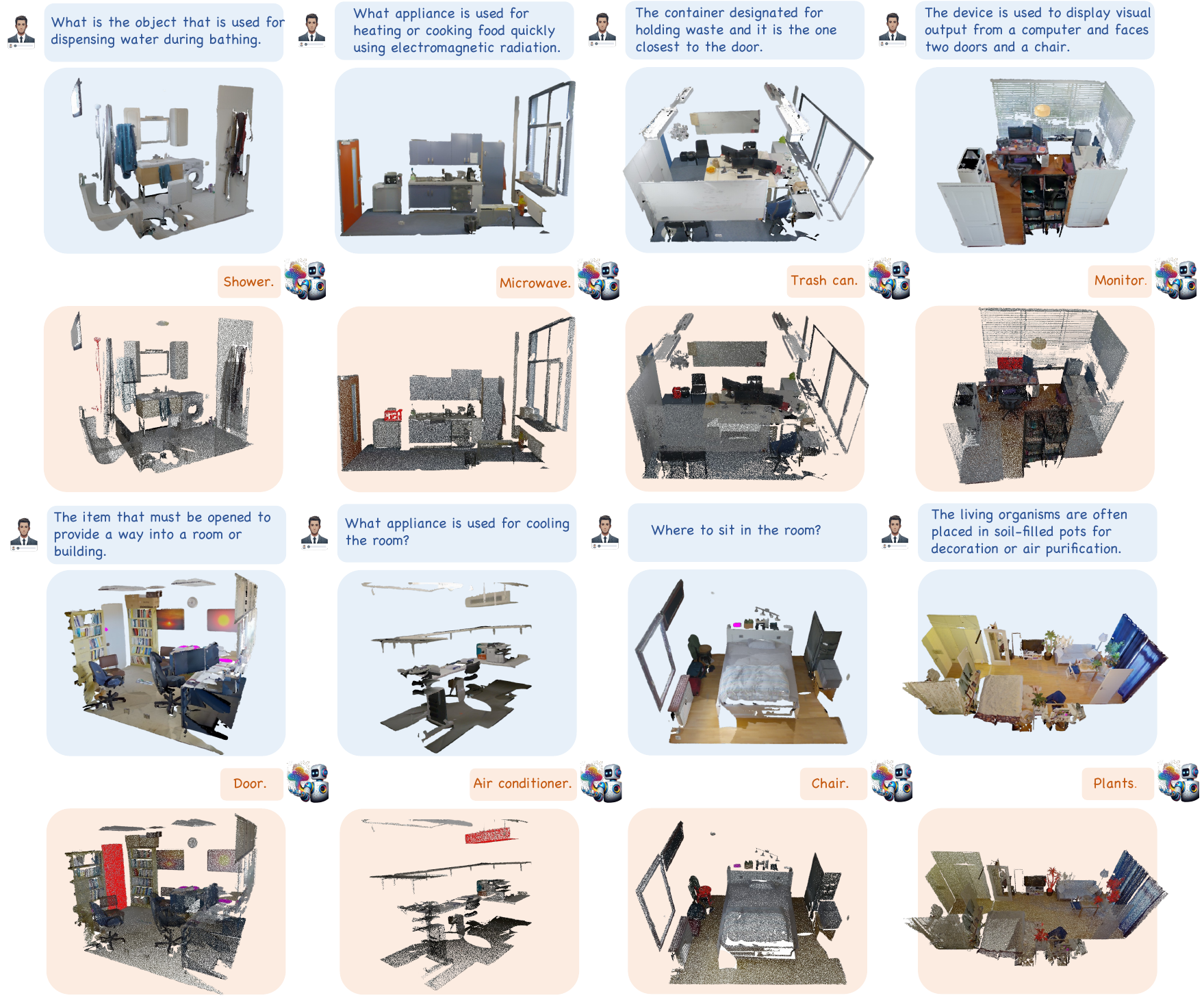}
	\vspace{-0.5cm}
	\caption{Qualitative results from val split of \ourdataset. \ours understand the human instruction and accurately segment the target object. We omitted the ``please output segmentation mask'' in the sentence for simplicity.} 
	\vspace{-0.6cm}
	\label{fig:visualization}
\end{figure}

As shown in \figurename~\ref{fig:visualization}, we provide some typical qualitative results from \ours on \ourdataset.
Given an implicit instruction, \eg, “\textit{the object that is used for
dispensing water during bathing}”,
\ours successfully infers the role of a shower.
In other scenes, the instructions include queries requiring extensive knowledge of the world and complex reasoning, like “\textit{What appliance is used for
heating or cooking food quickly using electromagnetic radiation.}”
\ours can still segment the microwave oven very well, 
which shows that \ours can make good use of the reasoning ability of LLM and provide high-quality masks.

\vspace{-4mm}
\section{Conclusion}
\vspace{-1mm}
In this work, we introduce \ours, an effective model supported by LLM for point-level reasoning and segmentation. Benefiting from the proposed Geometric Enhancer Module and Geometric-guided Feature Propagation,  \ours is adept at solving a variety of segmentation tasks in a unified framework. 
Additionally, we construct a comprehensive \ourdataset benchmark to bolster research area in segmentation via implicit and complex instructions, introducing more challenges to promote it closer to real-world applications. Through thorough experiments, \ours achieves promising results across multiple benchmarks.

\vspace{1mm}
\noindent\textbf{Limitations}
Although \ours demonstrates notable success in tasks driven by text prompts, its current framework cannot process non-textual prompts, such as boxes and points. Future developments will explore the adoption of a prompt encoder, inspired by SAM model~\cite{SAM}, to extend support for these formats.

\small{\noindent\textbf{Acknowledgements} We thank the anonymous reviewers for their constructive suggestions. Following their advice, we have incorporated diverse location and view descriptions into our \ourdataset. This work was partially supported by the National Research Foundation Singapore Competitive Research Program (CRP29-2022-0003).}


%
%
\bibliographystyle{splncs04}
\bibliography{main}

\begin{thebibliography}{10}
\providecommand{\url}[1]{\texttt{#1}}
\providecommand{\urlprefix}{URL }
\providecommand{\doi}[1]{https://doi.org/#1}

\bibitem{referit3d}
Achlioptas, P., Abdelreheem, A., Xia, F., Elhoseiny, M., Guibas, L.: Referit3d: Neural listeners for fine-grained 3d object identification in real-world scenes. In: ECCV (2020)

\bibitem{alayrac2022flamingo}
Alayrac, J.B., Donahue, J., Luc, P., Miech, A., Barr, I., Hasson, Y., Lenc, K., Mensch, A., Millican, K., Reynolds, M., et~al.: Flamingo: a visual language model for few-shot learning. In: NeurIPS (2022)

\bibitem{s3dis}
Armeni, I., Sener, O., Zamir, A.R., Jiang, H., Brilakis, I., Fischer, M., Savarese, S.: 3d semantic parsing of large-scale indoor spaces. In: CVPR (2016)

\bibitem{scanrefer}
Chen, D.Z., Chang, A.X., Nie{\ss}ner, M.: Scanrefer: 3d object localization in rgb-d scans using natural language. In: ECCV (2020)

\bibitem{vitadapter}
Chen, Z., Duan, Y., Wang, W., He, J., Lu, T., Dai, J., Qiao, Y.: Vision transformer adapter for dense predictions. In: ICLR (2023)

\bibitem{minkowski}
Choy, C., Gwak, J., Savarese, S.: 4d spatio-temporal convnets: Minkowski convolutional neural networks. In: CVPR (2019)

\bibitem{scannet}
Dai, A., Chang, A.X., Savva, M., Halber, M., Funkhouser, T., Nie{\ss}ner, M.: Scannet: Richly-annotated 3d reconstructions of indoor scenes. In: CVPR (2017)

\bibitem{MeViS}
Ding, H., Liu, C., He, S., Jiang, X., Loy, C.C.: {MeViS}: A large-scale benchmark for video segmentation with motion expressions. In: ICCV (2023)

\bibitem{MOSE}
Ding, H., Liu, C., He, S., Jiang, X., Torr, P.H., Bai, S.: {MOSE}: A new dataset for video object segmentation in complex scenes. In: ICCV (2023)

\bibitem{ding2021vision}
Ding, H., Liu, C., Wang, S., Jiang, X.: Vision-language transformer and query generation for referring segmentation. In: ICCV (2021)

\bibitem{vltpami}
Ding, H., Liu, C., Wang, S., Jiang, X.: {VLT}: Vision-language transformer and query generation for referring segmentation. IEEE TPAMI  (2023)

\bibitem{PLA}
Ding, R., Yang, J., Xue, C., Zhang, W., Bai, S., Qi, X.: Pla: Language-driven open-vocabulary 3d scene understanding. In: CVPR (2023)

\bibitem{maskclip}
Ding, Z., Wang, J., Tu, Z.: Open-vocabulary panoptic segmentation with maskclip. In: ICLR (2023)

\bibitem{imagebind}
Girdhar, R., El-Nouby, A., Liu, Z., Singh, M., Alwala, K.V., Joulin, A., Misra, I.: Imagebind: One embedding space to bind them all. In: CVPR (2023)

\bibitem{pointbind}
Guo, Z., Zhang, R., Zhu, X., Tang, Y., Ma, X., Han, J., Chen, K., Gao, P., Li, X., Li, H., et~al.: Point-bind \& point-llm: Aligning point cloud with multi-modality for 3d understanding, generation, and instruction following. arXiv preprint arXiv:2309.00615  (2023)

\bibitem{resnet}
He, K., Zhang, X., Ren, S., Sun, J.: Deep residual learning for image recognition. In: CVPR (2016)

\bibitem{DsHmp}
He, S., Ding, H.: Decoupling static and hierarchical motion perception for referring video segmentation. In: CVPR (2024)

\bibitem{RefMask3D}
He, S., Ding, H.: {RefMask3D}: Language-guided transformer for 3d referring segmentation. In: ACM MM (2024)

\bibitem{PAPFZ}
He, S., Jiang, X., Jiang, W., Ding, H.: Prototype adaption and projection for few-and zero-shot 3d point cloud semantic segmentation. IEEE TIP  (2023)

\bibitem{3dllm}
Hong, Y., Zhen, H., Chen, P., Zheng, S., Du, Y., Chen, Z., Gan, C.: 3d-llm: Injecting the 3d world into large language models. In: NeurIPS (2023)

\bibitem{LoRA}
Hu, E.J., Shen, Y., Wallis, P., Allen-Zhu, Z., Li, Y., Wang, S., Wang, L., Chen, W.: Lora: Low-rank adaptation of large language models. In: ICLR (2022)

\bibitem{TGNN}
Huang, P.H., Lee, H.H., Chen, H.T., Liu, T.L.: Text-guided graph neural networks for referring 3d instance segmentation. In: AAAI (2021)

\bibitem{butd-detr}
Jain, A., Gkanatsios, N., Mediratta, I., Fragkiadaki, K.: Bottom up top down detection transformers for language grounding in images and point clouds. In: ECCV (2022)

\bibitem{sceneverse}
Jia, B., Chen, Y., Yu, H., Wang, Y., Niu, X., Liu, T., Li, Q., Huang, S.: Sceneverse: Scaling 3d vision-language learning for grounded scene understanding. In: ECCV (2024)

\bibitem{pointgroup}
Jiang, L., Zhao, H., Shi, S., Liu, S., Fu, C.W., Jia, J.: Pointgroup: Dual-set point grouping for 3d instance segmentation. In: CVPR (2020)

\bibitem{SAM}
Kirillov, A., Mintun, E., Ravi, N., Mao, H., Rolland, C., Gustafson, L., Xiao, T., Whitehead, S., Berg, A.C., Lo, W.Y., et~al.: Segment anything. In: ICCV (2023)

\bibitem{oneformer3d}
Kolodiazhnyi, M., Vorontsova, A., Konushin, A., Rukhovich, D.: Oneformer3d: One transformer for unified point cloud segmentation. In: CVPR (2024)

\bibitem{LISA}
Lai, X., Tian, Z., Chen, Y., Li, Y., Yuan, Y., Liu, S., Jia, J.: Lisa: Reasoning segmentation via large language model. In: CVPR (2024)

\bibitem{MAFT}
Lai, X., Yuan, Y., Chu, R., Chen, Y., Hu, H., Jia, J.: Mask-attention-free transformer for 3d instance segmentation. In: ICCV (2023)

\bibitem{li2023otter}
Li, B., Zhang, Y., Chen, L., Wang, J., Yang, J., Liu, Z.: Otter: A multi-modal model with in-context instruction tuning. arXiv:2305.03726  (2023)

\bibitem{li2023blip}
Li, J., Li, D., Savarese, S., Hoi, S.: Blip-2: Bootstrapping language-image pre-training with frozen image encoders and large language models. In: NeurIPS (2023)

\bibitem{Sphinx}
Lin, Z., Liu, C., Zhang, R., Gao, P., Qiu, L., Xiao, H., Qiu, H., Lin, C., Shao, W., Chen, K., et~al.: Sphinx: The joint mixing of weights, tasks, and visual embeddings for multi-modal large language models. arXiv preprint arXiv:2311.07575  (2023)

\bibitem{GRES}
Liu, C., Ding, H., Jiang, X.: {GRES}: Generalized referring expression segmentation. In: CVPR (2023)

\bibitem{M3Att}
Liu, C., Ding, H., Zhang, Y., Jiang, X.: Multi-modal mutual attention and iterative interaction for referring image segmentation. IEEE TIP  (2023)

\bibitem{ISFP}
Liu, C., Jiang, X., Ding, H.: Instance-specific feature propagation for referring segmentation. IEEE TMM  (2022)

\bibitem{primitivenet}
Liu, C., Jiang, X., Ding, H.: Primitivenet: decomposing the global constraints for referring segmentation. Visual Intelligence  \textbf{2}(1), ~16 (2024)

\bibitem{liu2023visual}
Liu, H., Li, C., Wu, Q., Lee, Y.J.: Visual instruction tuning. In: NeurIPS (2023)

\bibitem{seal}
Liu, Y., Kong, L., Cen, J., Chen, R., Zhang, W., Pan, L., Chen, K., Liu, Z.: Segment any point cloud sequences by distilling vision foundation models. In: NeurIPS (2023)

\bibitem{loshchilov2017decoupled}
Loshchilov, I., Hutter, F.: Decoupled weight decay regularization. In: ICLR (2019)

\bibitem{open3dis}
Nguyen, P.D., Ngo, T.D., Gan, C., Kalogerakis, E., Tran, A., Pham, C., Nguyen, K.: Open3dis: Open-vocabulary 3d instance segmentation with 2d mask guidance. In: CVPR (2024)

\bibitem{park2022vision}
Park, N., Kim, S.: How do vision transformers work? arXiv preprint arXiv:2202.06709  (2022)

\bibitem{Openscene}
Peng, S., Genova, K., Jiang, C., Tagliasacchi, A., Pollefeys, M., Funkhouser, T., et~al.: Openscene: 3d scene understanding with open vocabularies. In: CVPR (2023)

\bibitem{peng2023kosmos}
Peng, Z., Wang, W., Dong, L., Hao, Y., Huang, S., Ma, S., Wei, F.: Kosmos-2: Grounding multimodal large language models to the world. In: ICLR (2024)

\bibitem{detgpt}
Pi, R., Gao, J., Diao, S., Pan, R., Dong, H., Zhang, J., Yao, L., Han, J., Xu, H., Zhang, L.K.T.: Detgpt: Detect what you need via reasoning. In: EMNLP (2023)

\bibitem{pointnet}
Qi, C.R., Su, H., Mo, K., Guibas, L.J.: Pointnet: Deep learning on point sets for 3d classification and segmentation. In: CVPR (2017)

\bibitem{pointnet++}
Qi, C.R., Yi, L., Su, H., Guibas, L.J.: Pointnet++: Deep hierarchical feature learning on point sets in a metric space. In: NeurIPS (2017)

\bibitem{GPT4Point}
Qi, Z., Fang, Y., Sun, Z., Wu, X., Wu, T., Wang, J., Lin, D., Zhao, H.: Gpt4point: A unified framework for point-language understanding and generation. In: CVPR (2024)

\bibitem{pointnext}
Qian, G., Li, Y., Peng, H., Mai, J., Hammoud, H., Elhoseiny, M., Ghanem, B.: Pointnext: Revisiting pointnet++ with improved training and scaling strategies. In: NeurIPS (2022)

\bibitem{X-RefSeg3D}
Qian, Z., Ma, Y., Ji, J., Sun, X.: X-refseg3d: Enhancing referring 3d instance segmentation via structured cross-modal graph neural networks. In: AAAI (2024)

\bibitem{GlaMM}
Rasheed, H., Maaz, M., Shaji, S., Shaker, A., Khan, S., Cholakkal, H., Anwer, R.M., Xing, E., Yang, M.H., Khan, F.S.: Glamm: Pixel grounding large multimodal model. In: CVPR (2024)

\bibitem{rasley2020deepspeed}
Rasley, J., Rajbhandari, S., Ruwase, O., He, Y.: Deepspeed: System optimizations enable training deep learning models with over 100 billion parameters. In: KDD (2020)

\bibitem{PixelLM}
Ren, Z., Huang, Z., Wei, Y., Zhao, Y., Fu, D., Feng, J., Jin, X.: Pixellm: Pixel reasoning with large multimodal model. In: CVPR (2024)

\bibitem{scannet200}
Rozenberszki, D., Litany, O., Dai, A.: Language-grounded indoor 3d semantic segmentation in the wild. In: ECCV (2022)

\bibitem{mask3d}
Schult, J., Engelmann, F., Hermans, A., Litany, O., Tang, S., Leibe, B.: Mask3d: Mask transformer for 3d semantic instance segmentation. In: ICRA (2023)

\bibitem{shuai2024survey}
Shuai, X., Ding, H., Ma, X., Tu, R., Jiang, Y.G., Tao, D.: A survey of multimodal-guided image editing with text-to-image diffusion models. arXiv preprint arXiv:2406.14555  (2024)

\bibitem{spformer}
Sun, J., Qing, C., Tan, J., Xu, X.: Superpoint transformer for 3d scene instance segmentation. In: AAAI (2023)

\bibitem{openmask3d}
Takmaz, A., Fedele, E., Sumner, R.W., Pollefeys, M., Tombari, F., Engelmann, F.: Openmask3d: Open-vocabulary 3d instance segmentation. In: NeurIPS (2023)

\bibitem{kpconv}
Thomas, H., Qi, C.R., Deschaud, J.E., Marcotegui, B., Goulette, F., Guibas, L.J.: Kpconv: Flexible and deformable convolution for point clouds. In: ICCV (2019)

\bibitem{llama}
Touvron, H., Lavril, T., Izacard, G., Martinet, X., Lachaux, M.A., Lacroix, T., Rozi{\`e}re, B., Goyal, N., Hambro, E., Azhar, F., et~al.: Llama: Open and efficient foundation language models. arXiv preprint arXiv:2302.13971  (2023)

\bibitem{wang2023octformer}
Wang, P.S.: Octformer: Octree-based transformers for 3d point clouds. In: SIGGRAPH (2023)

\bibitem{wang2023visionllm}
Wang, W., Chen, Z., Chen, X., Wu, J., Zhu, X., Zeng, G., Luo, P., Lu, T., Zhou, J., Qiao, Y., et~al.: Visionllm: Large language model is also an open-ended decoder for vision-centric tasks. In: NeurIPS (2023)

\bibitem{pvtv2}
Wang, W., Xie, E., Li, X., Fan, D.P., Song, K., Liang, D., Lu, T., Luo, P., Shao, L.: Pvt v2: Improved baselines with pyramid vision transformer. Computational Visual Media  (2022)

\bibitem{3D-STMN}
Wu, C., Ma, Y., Chen, Q., Wang, H., Luo, G., Ji, J., Sun, X.: 3d-stmn: Dependency-driven superpoint-text matching network for end-to-end 3d referring expression segmentation. In: AAAI (2024)

\bibitem{cvt}
Wu, H., Xiao, B., Codella, N., Liu, M., Dai, X., Yuan, L., Zhang, L.: Cvt: Introducing convolutions to vision transformers. In: ICCV (2021)

\bibitem{OpenVocabulary}
Wu, J., Li, X., Xu, S., Yuan, H., Ding, H., Yang, Y., Li, X., Zhang, J., Tong, Y., Jiang, X., Ghanem, B., Tao, D.: Towards open vocabulary learning: A survey. IEEE TPAMI  (2024)

\bibitem{pointtransformerv2}
Wu, X., Lao, Y., Jiang, L., Liu, X., Zhao, H.: Point transformer v2: Grouped vector attention and partition-based pooling. In: NeurIPS (2022)

\bibitem{EDA}
Wu, Y., Cheng, X., Zhang, R., Cheng, Z., Zhang, J.: Eda: Explicit text-decoupling and dense alignment for 3d visual grounding. In: CVPR (2023)

\bibitem{xiao2023position}
Xiao, Z., Zhang, W., Wang, T., Loy, C.C., Lin, D., Pang, J.: Position-guided point cloud panoptic segmentation transformer. arXiv preprint arXiv:2303.13509  (2023)

\bibitem{ODISE}
Xu, J., Liu, S., Vahdat, A., Byeon, W., Wang, X., De~Mello, S.: Open-vocabulary panoptic segmentation with text-to-image diffusion models. In: CVPR (2023)

\bibitem{pointllm}
Xu, R., Wang, X., Wang, T., Chen, Y., Pang, J., Lin, D.: Pointllm: Empowering large language models to understand point clouds. In: ECCV (2024)

\bibitem{RegionPLC}
Yang, J., Ding, R., Deng, W., Wang, Z., Qi, X.: Regionplc: Regional point-language contrastive learning for open-world 3d scene understanding. In: CVPR (2024)

\bibitem{swin3d}
Yang, Y.Q., Guo, Y.X., Xiong, J.Y., Liu, Y., Pan, H., Wang, P.S., Tong, X., Guo, B.: Swin3d: A pretrained transformer backbone for 3d indoor scene understanding. arXiv preprint arXiv:2304.06906  (2023)

\bibitem{ye2023mplug}
Ye, Q., Xu, H., Xu, G., Ye, J., Yan, M., Zhou, Y., Wang, J., Hu, A., Shi, P., Shi, Y., et~al.: mplug-owl: Modularization empowers large language models with multimodality. arXiv:2304.14178  (2023)

\bibitem{scannet++}
Yeshwanth, C., Liu, Y.C., Nie{\ss}ner, M., Dai, A.: Scannet++: A high-fidelity dataset of 3d indoor scenes. In: ICCV (2023)

\bibitem{Ferret}
You, H., Zhang, H., Gan, Z., Du, X., Zhang, B., Wang, Z., Cao, L., Chang, S.F., Yang, Y.: Ferret: Refer and ground anything anywhere at any granularity. In: ICLR (2024)

\bibitem{LLaVA-grounding}
Zhang, H., Li, H., Li, F., Ren, T., Zou, X., Liu, S., Huang, S., Gao, J., Zhang, L., Li, C., et~al.: Llava-grounding: Grounded visual chat with large multimodal models. arXiv preprint arXiv:2312.02949  (2023)

\bibitem{PMOSR}
Zhang, H., Ding, H.: Prototypical matching and open set rejection for zero-shot semantic segmentation. In: ICCV (2021)

\bibitem{zhang2023gpt4roi}
Zhang, S., Sun, P., Chen, S., Xiao, M., Shao, W., Zhang, W., Chen, K., Luo, P.: Gpt4roi: Instruction tuning large language model on region-of-interest. arXiv:2307.03601  (2023)

\bibitem{Multi3drefer}
Zhang, Y., Gong, Z., Chang, A.X.: Multi3drefer: Grounding text description to multiple 3d objects. In: ICCV (2023)

\bibitem{pointtransformerv1}
Zhao, H., Jiang, L., Jia, J., Torr, P.H., Koltun, V.: Point transformer. In: ICCV (2021)

\bibitem{uni3d}
Zhou, J., Wang, J., Ma, B., Liu, Y.S., Huang, T., Wang, X.: Uni3d: Exploring unified 3d representation at scale. In: ICLR (2024)

\bibitem{zhou2021panoptic}
Zhou, Z., Zhang, Y., Foroosh, H.: Panoptic-polarnet: Proposal-free lidar point cloud panoptic segmentation. In: CVPR (2021)

\bibitem{zhu2023minigpt}
Zhu, D., Chen, J., Shen, X., Li, X., Elhoseiny, M.: Minigpt-4: Enhancing vision-language understanding with advanced large language models. In: ICLR (2024)

\bibitem{3d-vista}
Zhu, Z., Ma, X., Chen, Y., Deng, Z., Huang, S., Li, Q.: 3d-vista: Pre-trained transformer for 3d vision and text alignment. In: ICCV (2023)

\end{thebibliography}
\end{document}